\documentclass[conference]{IEEEtran}
\IEEEoverridecommandlockouts
\usepackage{cite}
\usepackage{amsmath,amssymb,amsfonts}
\usepackage{algorithmic}
\usepackage{graphicx}
\usepackage{textcomp}
\usepackage{xcolor}
\usepackage{soul}

\usepackage{verbatim}

\def\BibTeX{{\rm B\kern-.05em{\sc i\kern-.025em b}\kern-.08em
    T\kern-.1667em\lower.7ex\hbox{E}\kern-.125emX}}

\usepackage{booktabs} % For formal tables
\usepackage{todonotes}
\usepackage{url}
\usepackage{amsthm}
\usepackage{xcolor}

\usepackage{caption}
\usepackage{subcaption}
\begin{document}
\bstctlcite{IEEEexample:BSTcontrol}
%---TITLE----

\title{Capturing Local and Global Patterns in Procedural Content Generation via Machine Learning}

%\title{Confronting the Challenge of Capturing Local and Global Patterns in Procedural Content Generation}
%\title{Capturing Local and Global Patterns in PCGML}
%Local vs. Global Patterns in PCGML}

%---AUTHORS----
%COG Is not double blind so we should add author names
\author{\IEEEauthorblockN{Vanessa Volz\textsuperscript{1}, Niels Justesen\textsuperscript{1}, Sam Snodgrass\textsuperscript{1}, Sahar Asadi\textsuperscript{2}, Sami Purmonen\textsuperscript{2},\\ Christoffer Holmg\r{a}rd\textsuperscript{1}, Julian Togelius\textsuperscript{1}, Sebastian Risi\textsuperscript{1}}
\IEEEauthorblockA{\textsuperscript{1}\textit{modl.ai}, \textsuperscript{2}\textit{King Digital Entertainment}\\
}
}

%\author{\IEEEauthorblockN{Anonymous Author(s)}
%}

%-------
\maketitle

%---ABSTRACT----
\begin{abstract}
%Procedural content generation approaches have shown promising results in a variety of different domains.
%julian This sentence does not add much
Recent \emph{procedural content generation via machine learning} (PCGML) methods allow learning from existing content to produce similar content automatically. While these approaches are able to generate content for different games (e.g.\ Super Mario Bros., DOOM, Zelda, and Kid Icarus), it is an open questions how well these approaches can capture large-scale visual patterns such as symmetry. In this paper, we propose match-three games as a domain to test PCGML algorithms regarding their ability to generate suitable patterns. We demonstrate that popular algorithm such as Generative Adversarial Networks struggle in this domain and propose adaptations to improve their performance. In particular we augment the neighborhood of a Markov Random Fields approach to not only take local but also symmetric positional information into account. We conduct several empirical tests including a user study that show the improvements achieved by the proposed modifications, and obtain promising results.
%generating such patterns and investigate the ability of algorithms such as generative and adversarial networks (GANs) to do so. Additionally, we propose a a modification to the Markov Random Field algorithm that allows it to more easily capture both types of patterns.\textcolor{orange}{ An extensive user study, with X Candy Crush level designers? was performed, showing that ....}.
\end{abstract}

%---KEYWORDS----

%---Julian starting from scratch
\section{Introduction}
\label{sec:introdcution}

One approach to generating game content is training a machine learning model on %a large number of 
 available examples, and then sampling from the trained model for new game content that is similar to what it was trained on. This approach, generally referred to as Procedural Content Generation via Machine Learning (PCGML)~\cite{Summerville2018a}, assumes that the machine learning algorithm will learn the appropriate invariants of the content it is trained on, so that the content sampled from the learned model retains the ``style'' of the content, while introducing variety. For example, if trained on platform game levels, you would want the model to have learned the maximal lengths of gaps it can generate to still obtain playable levels, but to also introduce new challenges and scenarios (e.g.\ combinations of gaps) not found in the original levels.

%Super Mario Bros. underground levels, you would want the model to have learned that underground levels place blocks at the top of a level
%platform game levels,you would want the model to have learned how to make the levels playable and look ``platformer-like,'' but you would also want the generated levels to include new challenges and scenarios not found in the original levels.

Given that PCGML is a relatively recent research topic, it is perhaps not surprising that little research has been done on which invariants of the training contents are learned, and how to make the system learn the most relevant ones. In this paper, we investigate which types of patterns are captured by models trained on a large set of levels, and methods for capturing more relevant patterns. We provide a taxonomy of such methods in this paper, in order to help structure research in this direction.
%To help structure this research direction, we provide a taxonomy of methods for making machine learning methods better capture patterns in game content.

To test our methods we use levels from the mobile casual puzzle game Candy Crush Saga (CCS). Casual puzzle games as a genre are well-suited to PCG in general and to PCGML in particular, given the huge demand for level content; e.g. more than six thousand levels have been made for CCS. Just like in many other game genres, the quality of the levels is crucial to keep the player engaged. Interestingly, it is also a game genre for which, to our knowledge, few PCG studies and no PCGML studies exist in the literature.

Our results suggest existing PCGML methods mostly focus on local patterns and are unable to reproduce global features such as symmetry. However, we find certain methodological advancements, such as incorporating symmetric neighbors into Markov Random Fields, improve the methods in this regard. While these results are promising, consistently generating levels with the same complex local and global patterns as in the existing CCS levels is an important future research direction.

%---BACKGROUND----
\section{Background}
\label{sec:background}
%---------
%\subsection{Procedural Content Generation via Machine Learning}

\subsection{Patterns in Games}
\label{sec:background-patterns}
The term \emph{patterns} has different connotations in different domains, but they generally describe regularities within a given object. In the context of frequent pattern mining, this is taken to mean sets of items, sub-sequences or sub-structures that occur multiple times. In the context of generation of abstract images, patterns have previously been classified as symmetry, repetition, and repetition with variation \cite{stanley2003taxonomy}. In games, the most prominent use of patterns is in \emph{game design patterns}~\cite{bjork2005patterns}. Game design patterns can take many shapes, including rather abstract patterns related to the overall game design, which we do not consider here. But there are also more fine-grained patterns, including various spatial features observed in games such as first-person shooters~\cite{hullett2010design} and platformers~\cite{dahlskog2012patterns}. The type of patterns we address in this paper are fine-grained and visible spatial patterns, in particular relations between tiles.

While humans seem to have an intuitive understanding of what a visual pattern is, it is challenging to provide a general formal definition. In addition, games are ultimately made for humans -- a formal definition that does not align with human perception thus would not be meaningful. Here, we choose to mainly rely on a user study to identify patterns accompanied by a simple context-specific description of the concept.

In particular, we focus on Candy Crush Saga (CCS), which is a free-to-play match-three puzzle game released by King\footnote{\url{https://king.com/}} in 2012 and has since enjoyed major success as one of the top mobile casual games\footnote{data source: appannie.com}. In CCS, three or more candies (tiles) can be matched horizontally or vertically with neighbouring candies of the same colour. When matched, candies disappear from the board. If there are no obstructing items, this causes existing candies to fall down and fill the resulting gaps, and new candies with random colours to be spawned (usually at the top of the board). The game introduces various constraints, obstacles and objectives that together define each level and thus create puzzles of varying difficulty.  
%
%The game board in CCS is a grid of $9\times9$ cells that can be filled with game elements or remain empty. The objective of each level is defined based on its game mode. There are six different game modes in CCS. For example, one of the game modes is to match tiles on certain predefined tiles. These tiles are covered by jelly visualized by a white highlight on the tiles. To fulfill the objective of a jelly level, all tiles covered with jelly need to be cleared within a limited number of moves specified for that level. Through the combination of tiles with various effects, CCS can exhibit a large variety in gameplay.
%
In the context of CCS we use the terms:
\begin{itemize}
    \item \textbf{Global pattern}: A pattern that can only be identified by looking at spatial structure of \textbf{all} elements on the board.
    \item \textbf{Local pattern}: A pattern that can be identified in a \textbf{small area} of the board.
\end{itemize}

%--- Testing single column figures for taking less space
\begin{figure}
     \centering
     \begin{subfigure}[b]{0.4\columnwidth}
         \centering
         \includegraphics[width=\textwidth]{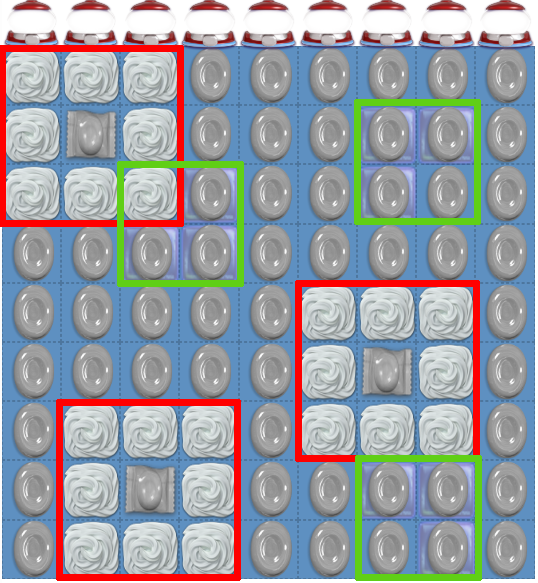}
         \caption{Local patterns}
         \label{fig:ex1}
     \end{subfigure}
     \hspace{.05\columnwidth}
     \begin{subfigure}[b]{0.4\columnwidth}
         \centering
         \includegraphics[width=\textwidth]{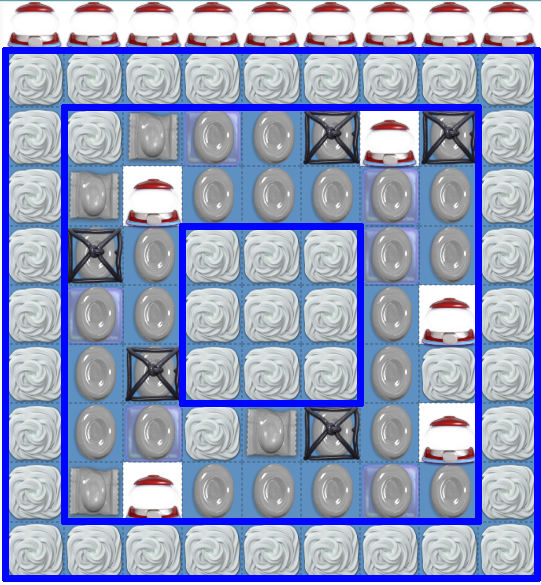}
         \caption{Global pattern}
         \label{fig:ex2}
     \end{subfigure}
     
    % \begin{subfigure}[b]{0.4\columnwidth}
    %     \centering
    %     \includegraphics[width=\textwidth]{pics/examples/examples-2-design_m.png}
    %     \caption{example1}
    %     \label{fig:ex3}
    %\end{subfigure}
    %\hspace{.05\columnwidth}
    %\begin{subfigure}[b]{0.4\columnwidth}
    %    \centering
    %    \includegraphics[width=\textwidth]{pics/examples/examples-3-game.png}
    %    \caption{example1}
    %    \label{fig:ex4}
    %\end{subfigure}
        \caption{Illustrative examples of patterns in synthetic CSS levels.}
        \label{fig:examples}
\end{figure}

CCS levels often exhibit global patterns commonly considered aesthetically pleasing or interesting to the human eye. Examples include placing items on the board in recognisable shapes or in symmetric arrangements (cf. Figure~\ref{fig:ex2}). 
%A level created to visualise a global pattern is depicted in Figure~\ref{fig:ex2}. 
There are also levels which repeat smaller, local patterns. A common example is that candies with additional beneficial effects are enclosed by obstructing items, making them harder to use (Figure~\ref{fig:ex1}). % In addition, as the different items influence other items with varying scale (some only affect neighbours, while others affect the whole board), it stands to reason that the mechanics thus also contribute to the existence of different patterns. An example of such a pattern is special candies (which act as explosives that can free larger areas), are locked between items that make it difficult to reach them.
Levels can display both local and global patterns.

The global aspect of CCS is different from standard applications of PCGML, such as Mario, where local tile neighbourhoods define the primary game elements (gaps, pipes, enemy groups). For this reason, we propose it as a test domain for PCGML algorithms to assess their ability to replicate different types of patterns, including local and global ones.

\subsection{Patterns in PCGML}
\label{sec:background-patterns-existing}

%Employing machine learning models tries to address one of the central issues of PCG, which is the need for an explicit and formal expression of the quality of generated content. Addressing this requirement is often difficult in the domain of games, as well as related subjective domains in arts and entertainment. Furthermore, these approaches are independent of the specific type of content generated, and are thus generally applicable.
%However, machine learning based generators and evaluators rely on some notion of similarity of the content. While parts of this similarity measure is often extracted from the training data (e.g.\ co-occurrence), it is still heavily restricted and biased by the approach chosen to capture similarity. This becomes an issue if this approach is not able to represent certain features that are important in the domain.

%Machine learning can be used in different ways within procedural content generation.
Algorithms performing Procedural Content Generation via Machine Learning (PCGML)~\cite{Summerville2018a}) directly employ machine learning models as content generators for games. 
%\begin{itemize}
%    \item as means to evaluate content generated by e.g.\ a search algorithm in search-based PCG~\cite{summerville2017understanding,guzdial2016deep,marino2015empirical}. This includes use of machine learning models for experience prediction in Experience-driven PCG~\cite{yannakakis2011experience}.
%    \item directly as content generators, which is known as \emph{procedural content generation via machine learning} PCGML~\cite{Summerville2018a}). This can be done through sampling the content generator randomly, or through searching for good model inputs, thereby searching the space created by the learned model~\cite{Volz2018}.
%\end{itemize}
%Many PCGML approaches for tile-based games express similarity by analysing patterns of adjacent tiles~\cite{lucas2019tile,snodgrass2015hierarchical}. While this approach has been demonstrated to work well for tasks such as Mario level generation~\cite{snodgrass2016learning, karth2017wavefunctioncollapse, jain2016autoencoders}, global metrics such as symmetry cannot be captured in this context. However, symmetry and other global patterns have been widely used to express the quality of board and puzzle games~\cite{browne2013metrics}.
%\todo[inline]{Bayesian approaches vs. others? (frequentist, belief functions, etc)}
At its core, PCGML is applied to learn invariant patterns from a set of examples. The type and scale of patterns captured is largely determined by the underlying machine learning approach, the training data, and its representation. The majority of existing PCGML techniques represent the data and patterns as local relationships between small scale elements either sequentially~\cite{summerville2016mariostring,snodgrass2016learning,dahlskog2014linear,hoover2015composing} or as local neighborhoods~\cite{snodgrass2016learning,karth2017wavefunctioncollapse,Volz2018,lucas2019tile,jain2016autoencoders}. These approaches have been shown to capture local patterns and small structures from the provided training data (e.g.\ pipes in \textit{Super Mario Bros.}~\cite{summerville2016mariostring,Volz2018}, or pits with water in a generic platform game~\cite{karth2017wavefunctioncollapse}). However, the ability of these approaches to reproduce and model larger scale and global patterns is typically not explored, and instead, computational metrics (e.g.\ linearity and leniency~\cite{snodgrass2016learning,summerville2016mariostring}) are used as proxies for global patterns. In this paper, we are interested in exploring the performance of these local models in domains where global patterns are a major aspect of the game.

A variety of extensions to local models have been developed to better capture larger scale patterns. Some have explored the idea of defining patterns explicitly either through clustering of, e.g., $3 \times 3 $ tile structres from levels~\cite{snodgrass2015hierarchical} or through training models on expert labeled patterns~\cite{Guzdial2018}. Others modified the generation step of the models to adhere to defined objectives and constraints~\cite{summerville2015mcmcts,snodgrass2016controllable}. Lastly, some have tried capturing larger patterns by modelling levels at multiple resolutions or employing hierarchical models~\cite{snodgrass2015hierarchical,gutierrez2020generative, summerville2015sampling,guzdial2016game}. While these extensions capture particular larger patterns or features, they are limited in what they each can capture; each extension was developed and tested with regard to specific patterns or features (e.g., $4\times4$ tile structures~\cite{snodgrass2015hierarchical}, item and enemy distributions~\cite{summerville2015mcmcts}, playability~\cite{snodgrass2016controllable}). 
%Instead, here we propose an overview of possible extensions with varying degrees of required specifications. %  In our work, we propose a framework for categorising extensions to local models, discuss different levels of generality for such extensions, and test the results of applying several such extensions.

%In addition to the local models and their extensions above, 
A number of PCGML techniques have been proposed that try to learn from the entirety of a level (or large sections) at once. 
Lee et al.~\cite{lee2016predicting} train Deep Convolutional Neural Networks on full \textit{StarCraft 2} maps and use them to predict resource placement. Their model is trained on full maps, but is meant to learn a specific feature and not a model of a full level. Autoencoders have been used for modeling and blending large sections from platforming levels~\cite{sarkar19controllable}, but they also rely on level features, such as linearity and density, as proxies for the global patterns. Shaker et al.~\cite{shaker2014alone} present a non-negative matrix factorization approach that splits the levels into different layers corresponding to level structure and object placements. %structure, item placements, and enemy placements; and separate sub-models are used to capture the patterns within each layer. 
However, this approach is evaluated with level features as proxies to evaluate whether patterns are reproduced. %Although we test our proposed extensions for better capturing global patterns on local models, they can be used to augment these larger scale models as well.
\section{Towards Pattern-Aware PCGML}
\label{sec:pattern-aware-adaptation}
As detailed above, most existing PCGML algorithms  focuse on capturing local patterns in game content. However, as we demonstrate in Section~\ref{sec:results}, this does not necessarily ensure generating global patterns we see in typical CCS levels (Figure~\ref{fig:vis}), such as symmetry. In the following, we present an overview of adaptations to improve existing PCGML methods in this regard. Three main approaches are identified, each of which relies on varying degrees of domain knowledge.%, that could be gathered through experts or user surveys.

%\begin{figure}
%    \centering
%    \includegraphics[width=1.0\columnwidth]{pics/overview.pdf}
%    \caption{Approaches to encourage the creation of particular types of patterns typically focus one of three aspects in the generation pipeline. One strategy is to (a) enrich the data used for training with particular types of patterns or augmenting it with additional class label. Another strategy (b) focuses on augmenting the algorithms themselves, for example through particular types of symmetric activation functions. The third approach (c) encourages specific types of patterns by manipulating the already generated data. One such manipulation could be to remove solutions with patterns that are not e.g.\ symmetric or to force symmetry by mirroring one half of the constructed artefact.}
%    \label{fig:overview}
%\end{figure}

%-------
\subsection{Enriching the Data} %previously Enrich Data
\label{sec:enrich-data}

One way to encourage the generation of specific types of patterns is to enrich the data used for training. % (Figure~\ref{fig:overview}a).
An \textbf{explicit} way of doing so is by labelling each example with the type of global pattern displayed. Conditional GANs~\cite{mirza2014conditional}, for example, are well suited to handle this kind of data. %To reduce model complexity, 
Models can also be trained on each class separately. Both options aim to strengthen the signals around global patterns that are present in the data. We investigate the latter option in this paper (Section~\ref{sec:rq}).

A \textbf{data-driven} approach to enrich data could be to identify class labels automatically using unsupervised learning.  %such as clustering. 
If limited labeled data is available, a corresponding approach with a classifier that encodes a \textbf{learned bias} is also possible.
%-------
\subsection{Augmenting the Algorithm} %previously Add Input
\label{sec:add-input}
Many of the algorithms discussed in Section~\ref{sec:background-patterns-existing} only take into account neighbouring patterns. %One strategy to improve local models is to \textit{add inputs} for the algorithm to process. 
Apart from enriching the training data, one strategy to improve a local model is to augment the algorithm to ensure it focuses on the desired patterns. % (Figure~\ref{fig:overview}b).
If domain knowledge is available, this can be done \textbf{explicitly} by %adding inputs required to detect desired patterns. 
modifying the structure of the model to detect the desired patterns. For example, to generate outputs with symmetry, we can add positions that should be mirrored at a given position as input to the algorithm. %We can then add  feed the algorithm elements that we would expect to see mirrored in addition to local neighbors.
We test this approach using MRFs as a baseline algorithm.  %(Section~\ref{sec:mrf}).

Another approach is to feed measures describing desired features (e.g.\ a symmetry score) to the algorithm, so that recognising the fitness of an individual based on desired features is facilitated. This can be done e.g.\ by giving additional inputs to a GAN's discriminator. While these approaches are likely good at producing content exhibiting the desired features, they are heavily reliant on domain knowledge and the ability to characterise the features numerically.

Instead of relying on domain knowledge, another approach is to ensure that the input at least allows the algorithm to make connections between items at the scale of the desired global pattern. Examples of this \textbf{data-driven} approach could be adding a fully-connected layer as the first layer of the discriminator in a GAN. Another promising approach is adding the position of each input (e.g.\ as coordinates) to the input of a neural network. This has been explored before in the context of CPPNs~\cite{stanley:gpem2007}, GANs~\cite{uber:coordconv}, and their combination~\cite{schrum2020cppn2gan}.

Between these two extremes lie approaches with a \textbf{learned bias}. Such a bias can be learned through labeled samples or adversarial training and then given to the model as an additional input. An example would be attention layers for a neural network as proposed in \cite{pmlr-v97-zhang19d}.

%-------
\subsection{Filtering the Solutions} %previously Filter Solutions
\label{sec:filter-solutions}
A third approach is filtering  solutions, which is most straightforward if done \textbf{explicitly}, but it is conceivable to learn desired patterns and ways to identify them. Filtering can be executed at different times during the training process. Before training would mean creating a representation that only encodes solutions with the desired global patterns. In case of symmetry, only half of the level could be generated and automatically mirrored to construct the final level. Repairing solutions to adhere to patterns (e.g.\ through mirroring) is also possible during or after training. A further option for filtering after training is applying a search algorithm to the space of generated content, e.g.\ latent vector evolution for GANs~\cite{Volz2018}.

%\subsection{Adaptation of existing methods for pattern-aware PCGML}
%\label{sec:pattern-aware-adaptation}
%\begin{itemize}
%    \item Change info input for pattern discovery
%    \begin{itemize}
%        \item No bias: OneGAN (fully connected layer), Coord-Conv
%        \item Trained bias: Attention layers
%        \item Conscious bias: mrf-global, feed features to GAN
%    \end{itemize}
%    \item Separate training data according to patterns
%    \begin{itemize}
%        \item separate symmetry patterns (Hypothesis: faces work well because all have the same type of symmetry)
%    \end{itemize}
%    \item Force specific patterns
%    \begin{itemize}
%        \item Latent Variable search for specific patterns
%        \item repair to force patterns (e.g. mirroring)
%        \item representation constraints (at different points of the training process)
%    \end{itemize}
%\end{itemize}

%---EXPERIMENTAL SETUP------
\section{Experimental Setup}
\label{sec:experimental-setup}

%\subsubsection{Patterns in Candy Crush}

%\begin{itemize}
%    \item local
%    \begin{itemize}
%        \item areas of jelly often interconnected
%        \item special candies locked inside blocker areas
%        \item blockers near exit areas
%    \end{itemize}
%    \item global
%    \begin{itemize}
%        \item general co-occurrence (types of blocks with types of special candy etc)
%        \item symmetry
%        \item more free tiles on the top
%    \end{itemize}
%    \item playstyle
%    \begin{itemize}
%        \item level archetypes (explosive, sniper, etc)
%    \end{itemize}
%\end{itemize}

%\subsection{Example results}
%User study (pair-wise comparison of different approaches with around 10 designers; is there a difference between experts and non-experts? maybe also player), bot testing, ... 
%Quality diversity with different approaches (expressivity) - potential metrics (earth mover distance) 

In this paper, we evaluate state-of-the-art PCGML methods with regard to how well they are able to capture global and local patterns in a game. We extend this evaluation to several adaptations discussed in Section~\ref{sec:pattern-aware-adaptation}. We apply two popular PCGML techniques, namely Markov Random Fields (MRF)~\cite{snodgrass2016learning} and Generative Adversarial Networks (GAN)~\cite{Volz2018,gutierrez2020generative, torrado2019bootstrapping} to CCS level generation. As argued in Section~\ref{sec:background-patterns}, CCS levels exhibit several patterns at different scales. This review of the state-of-the-art is not exhaustive, and only intended as an illustration of the potential shortcomings of popular methods on games like CCS.

In the following, we describe our experimental setup; including the representation we use for CCS levels and the algorithms employed. % and the pre-processing to transform them into a suitable format to use as input to our PCGML algorithms.
%We present the specifications of our experimental setup and the algorithms employed.
%We then discuss how we evaluate the generators and formulate the research questions we intend to answer. 

%---------
\subsection{CCS Level Representation}
\label{sec:pcgml-candycrush}
CCS contains approximately $80$ game elements with different characteristics. Some levels rely on unique mechanics or game elements, which makes them difficult to replicate. For this reason, with the help of experts, we selected a subset of published CCS levels that are more homogeneous. We selected only levels from a specific game mode (\emph{Jelly}\footnote{\url{https://candycrush.fandom.com/wiki/Jelly_levels}}) and discarded levels containing complex dynamic elements such as \emph{frogs} and \emph{conveyor belts}, resulting in the $504$ levels we used in our experiments. There are still $51$ unique items present in the reduced set of levels. Some items can be stacked on the same cell in the board, and we found $789$ unique item stacks.

Most state-of-the-art PCGML approaches for tile-based content generation (Section~\ref{sec:background-patterns-existing}) are not equipped to scale up to this number of different tiles without incurring excessive computational costs. % or at least reach unexplored territory.
%In order to achieve a scale more commonly encountered in previous work, w
Thus, we introduce an abstract representation for CCS levels to reduce the representation complexity based on domain knowledge:
\begin{itemize}
    \item \textsc{Shape}: Indicates which game board cell are non-void.% (i.e., can contain the following elements).%Void tiles are displayed as transparent and indicate the borders of the game board.
    \item \textsc{Regular}: The six types of regular candy that can be matched with other candies of the same color.% indicated in different colours
    \item \textsc{Special}: Match-able items with additional effects
    \item \textsc{Block}: Items that obstructs matches by occupying a cell
    \item \textsc{Jelly}: Items indicating cells where matches need to occur to win the level in \emph{Jelly} game mode.
    \item \textsc{Lock}: Items that obstructs matches by restricting movement of items in the same cell
\end{itemize}

\noindent %The game board in CCS is a grid of size $9\times9$. % (tiles can be empty). 
%In our experiments, we defined $6$ categories. Thus, 
With the six categories mentioned above, each levels is represented as   a matrix with dimensions  $9\times9\times6$. We use a binary encoding to represent the occurrence of an item category in a given cell. However, in order to be able to ensure the validity of the generated CCS levels, we introduce the following  post-processing method for the GAN-based approaches.%\footnote{This step is not required for MRF algorithms as it generates levels only from valid item stacks.}

%\begin{itemize}
The first four layers cannot coexist in the same cell. The choice for each cell is determined by selecting the layer with highest value.\footnote{In case of ties, the order decides which layer is selected} %Which item is placed there is decided based on whichever layer has the highest value.
    However, if none of the values is higher than a threshold ($0.5$ in this case), the cell is indicated empty.
Further, only allow locks to be placed on cells that are not void or empty and jelly is only placed on cells that are not void.
%\end{itemize}

Additionally, in order to keep the complexity of the level representation low, we add the following mandatory post-processing steps to all levels in this experiment, including original ones: 
%\begin{itemize}
(1) Candies are spawned through candy cannons. Candy cannons are not included in this representation, so they are automatically placed above non-void cells in order to ensure that new candies are dropped. % as the tiles get cleared by matching same color candies.
(2) Additional, complex dynamic elements such as portals and special candy cannons are removed from the game to avoid introducing unnecessary complexity.
(3) As this paper focuses on visual patterns, we do not include further game meta-data such as the number of available moves in the representation.
%\end{itemize}

%A depiction of the representation can be seen in Figure~\ref{fig:rep}.

%\begin{figure}
%    \centering
%    \includegraphics[width=0.6\columnwidth]{pics/new-rep.png}
%    \caption{Caption}
%    \label{fig:rep}
%\end{figure}

%\begin{figure}
%    \centering
%    \includegraphics[width=0.4\textwidth]{pics/new-vis.png}
%    \caption{Caption}
%    \label{fig:my_label_two}
%\end{figure}

%---------
\subsection{Algorithms}
\label{sec:pcgml-algorithm}
This section introduces two baseline algorithms and our proposed extensions to them. We chose Markov Random Fields because they learn strictly local patterns, and GANs because they should be able to capture global patterns, as demonstrated in their application to face generation.

\begin{figure}
    \centering
    \includegraphics[width=.9\columnwidth]{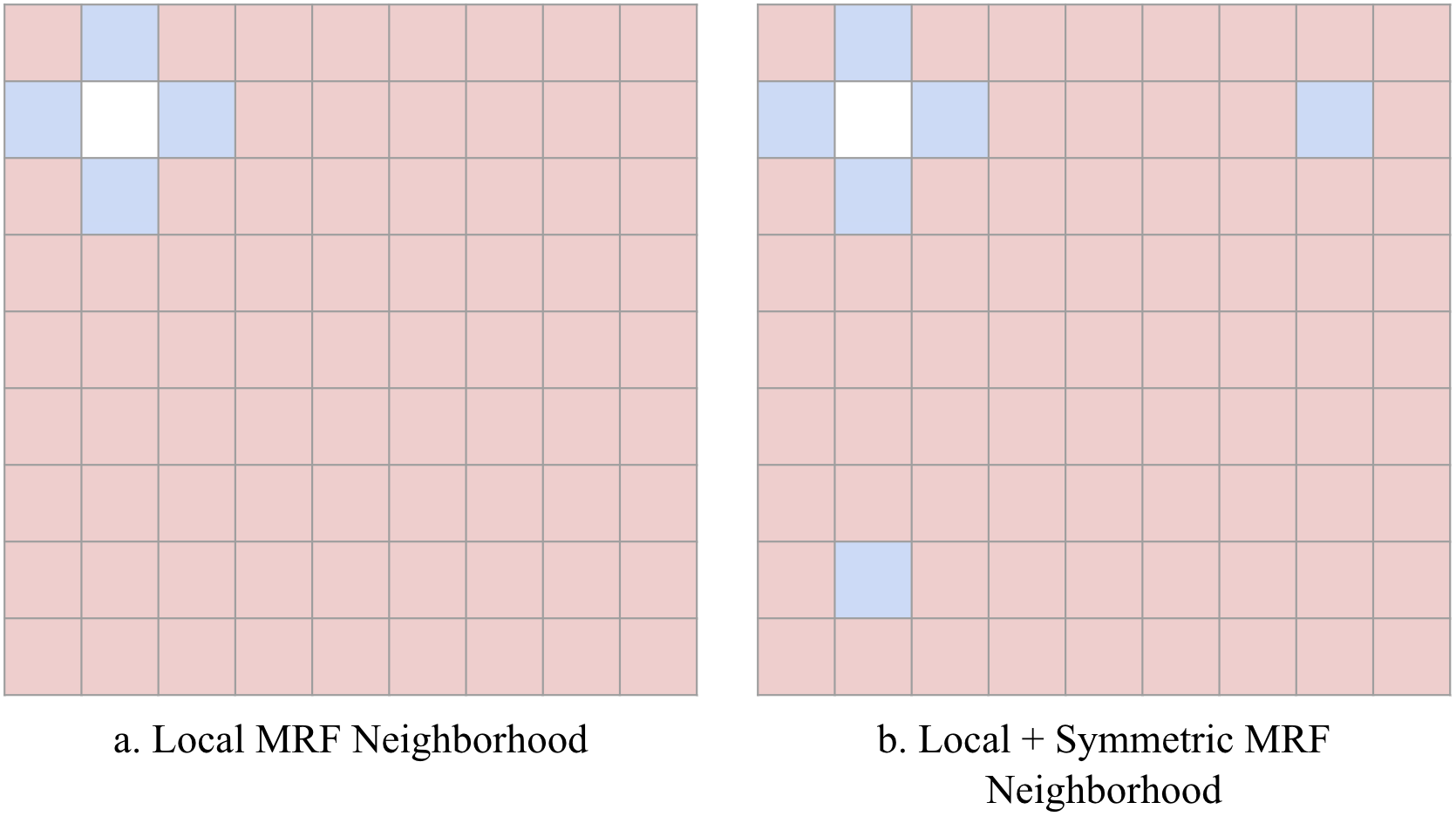}
    \caption{The neighbors used by our local Markov Random Field approach (left), and by our global MRF approach that includes vertically and horizontally symmetric positions (right). White indicates the current position, blue indicates positions the current is conditioned on, and red indicates independent positions.}
    \label{fig:MRF}
\end{figure}

\textbf{Markov Random Fields} (MRFs)~\cite{clifford1990markov} are undirected graphical models that have been used extensively in texture synthesis and repair~\cite{levina2006texture,hassner1981use}. MRFs model relationships between neighbouring positions (e.g., connected nodes in graphs or nearby pixels in textures). We build on previous work on level generation that used MRFs~\cite{snodgrass2016learning}, in which the authors represent a level using a grid of tile types corresponding to different level elements (e.g., solid objects, empty space, items, etc.), and treat the four surrounding positions in the grid as the neighbourhood of the MRF. This setup allowed the MRF to learn a conditional probability distribution (CPD) describing the probability of tile types at a position given the tile types of the neighbouring positions (Figure~\ref{fig:MRF}a).  %, and corresponds to the CPD $P(S_{x,y} | S_{x-1,y}, S_{x+1,y}, S_{x,y-1}, S_{x,y+1})$. 
We replicate this setup for our \textbf{local MRF}. To use this model with CCS,  the six-layer representation is collapsed into a single-layer representation where the value at a position is the concatenation of values in each layer. In this way,  each possible combination of values for the six layers is treated as a distinct tile type. For the \textbf{global MRF model}, the neighbourhood of the local MRF is modified  by adding the vertically and horizontally symmetric positions relative to the current position to the neighbourhood. Figure~\ref{fig:MRF}b shows an example of this modified neighbourhood. By including non-local positions, we hypothesize the MRF will better capture global patterns such as symmetry. Other MRF extensions have tested longer-range~\cite{li2008sparse,sun2011learning} and adaptive neighbourhoods~\cite{suwanwimolkul2018adaptive}; but to our knowledge, this is the first time MRFs have been extended with symmetric neighbors.

\textbf{Generative Adversarial Networks (GANs)} are a class of unsupervised machine learning methods where two neural networks, a generator and a discriminator, are trained in an adversarial setting~\cite{goodfellow2014generative}. The generator attempts to generate realistic data examples and the discriminator has to judge whether data examples come from the generator or a dataset of real examples. The Deep Convolutional GAN (DCGAN)~\cite{radford2015unsupervised} applies convolution and deconvolution to the discriminator and generator networks, enhancing the results on image data. We build on \textbf{MarioGAN}~\cite{Volz2018}, a DCGAN trained on levels from Super Mario Bros. %will improve the results on CCS levels. This  
The input data for MarioGAN was structured as $32 \times 32$ patches with one channel for each tile type (air, ground, etc.). In contrast to MarioGAN, we use $3\times3$ filters instead of $4 \times 4$ and produce $9\times9$ patches, which in the case of CCS is the entire level. All other hyper-parameters are kept the same and we train each model for $5,000$ epochs. We hypothesise that modifying the discriminator to use $9 \times 9$ filters would enable the discriminator to learn kernels that capture global patterns. Ideally, two streams of convolution could be employed, one with $9 \times 9$ filters for global patterns and one with $3 \times 3$ filters for local patterns. %It is, however, not clear how deconvolution in the generator can be modified to improve the production of global patterns. 
For simplicity, we employ just one stream of $9 \times 9$ filters in the discriminator, in the variant we call \textbf{GlobalGAN}. 

%--------
\subsection{Evaluation}
\label{sec:pgml-evaluation}
Our evaluation focuses on identifying whether visual patterns present in the training data are found in generated levels. In particular,  symmetry is evaluated as an example of a recognisable global pattern. % the ability of a generator to produce global patterns by analysing the symmetry of the generated levels. %measuring symmetry in the produced levels. %One specific aspect that we are going to focus on to evaluate a generator's ability to reproduce global patterns is the symmetry of the produced levels. 
We focus on symmetry as it is a common global pattern and is identifiable both visually and computationally. Still, as human perception can diverge from computational measures, we also conduct a user study.
%Because some patterns are difficult to define and evaluate quantitatively, we also conducted a user study among experts. %In the following, we give more details on the evaluations that we employ, namely anecdotal visual analysis, computational measures, and the user study.

%--------
\subsubsection{Visual analysis}
\label{sec:pcgml-visual-analysis}
As we are targeting aesthetic patterns, a visual inspection of samples of generated levels is suitable in this case. Here, we use it to manually pre-screen promising generators to be evaluated further in the user study.

%-------
\subsubsection{Computational Measures}
\label{sec:pcgml-computational-measures}
Horizontal, vertical, and diagonal line symmetry (also known as reflection or mirror symmetry) are measured. These measures are chosen because they were deemed important by level designers. Symmetry scores correspond to the ratio of positions on the board that are \emph{identical} to one of their symmetric counterparts. For the diagonal symmetry score,  both diagonal symmetry lines are considered at the same time, such that a position has to be identical to just one of the symmetric counterparts to count towards the score. Positions are considered to be identical if all layers are equal (see Section \ref{sec:pcgml-candycrush}). %A level that is perfectly horizontally symmetric means that then entire level is mirrored across the horizontal line. 
Our data set of CCS levels has a median of $100\%$ vertical symmetry and $55.6\%$ horizontal symmetry. 

% TODO: How do we refer to the data set. Has it been introdcued yet?
% TODO: Check that symmetry scores are the same across the entire paper.
% TODO: Check that we are consistent with horizontal and vertical symmetry

%-------
\subsubsection{User study}
\label{sec:pcgml-userstudy}
We conducted a user study among experts to identify and evaluate patterns that are not measurable computationally. In the following, we describe the design of the questionnaire distributed among King employees.\footnote{The full questionnaire can be accessed at\\ \url{https://forms.gle/nwAyPqckAZiEvBbr7}.}

The participants were presented with a short explanation and several examples of what global and local patterns mean in the context of this study (cf. Section~\ref{sec:background-patterns}). All levels were visualised using the original CCS sprites (see Figure~\ref{fig:vis}). Based on this definition, participants were asked the questions below:
%The generated levels for this study were synthetic but visualised in the same way as actual CCS levels.
%After a short explanation of local and global patterns in the context of CCS based on examples in Figure~\ref{fig:examples}, the questionnaire contains the following questions / tasks:
%
\begin{enumerate}
    \item Indicate all levels that contain a global pattern (checkboxes next to level images).
    \item Indicate all levels that contain one or more local patterns (checkboxes next to level images).
    \item Which level looks less like a CCS level? (pairwise comparison of two level images)
    %\item Which level would you rather play? (Pairwise comparison of two level images)
    \item Questions about experience of the participant in level design and game mechanics (multiple choice.)
\end{enumerate}

%The participants were presented with a short explanation and several examples of what global and local patterns mean in this task, in the context of CCS-type levels. The examples were illustrated using a game design visualisation tool that removes colours from the candies to highlight the underlying structures (Figure~\ref{fig:examples}). Based on this definition, participants were asked to identify which levels contain local and/or global patterns. The generated levels for this study were synthetic but visualised in the same way as actual CCS levels.  

%Decisions
%\begin{itemize}
%    \item 3 levels (best, worst, median according to symmetry score) for each generator
%    \item baselines vary for each section of test (see each level the same number of times)
%    \item total of 27 questions
%\end{itemize}

%\todo[inline]{add the above bullet points to text if needed}

%The user is presented with two levels (or alternatively, two sets of levels). They then have to pick between these two options for a set of questions. This process repeats for several sets of options

%\begin{enumerate}
%    \item Please indicate your level of expertise with Candy Crush Saga. (1. never heard of it, 2. haven't played, but played similar games, 3. have played - indicate level). Checkbox for [] I am a level designer.
%    \item Which of the levels looks more like a typical Candy Crush level?
%    \item Which of the levels would you think plays more like a typical Candy Crush level?
%    \item Which of the levels fits the design guidelines better?
%\end{enumerate}

%\todo[inline]{some answers incomplete?}

%------
\subsection{Research Questions}
\label{sec:rq}
We specifically want to investigate two types of adaptation suggested in Section~\ref{sec:pattern-aware-adaptation}, namely \emph{explicit algorithm augmentation} and \emph{explicit data enrichment}.

\paragraph*{RQ1}
Does extending the neighbourhood of an MRF improve its ability to produce global patterns? To answer this question, we compare our extended MRF (GlobalMRF) with the standard MRF (LocalMRF).

\paragraph*{RQ2}
Does training a GAN on only a single type of global pattern improve its ability to produce this pattern? We compare generators trained with the GlobalGAN architecture - one on the full set of levels (GlobalGAN), one on only vertically symmetric ones (GlobalGAN-vert). 

%We chose to use the GlobalGAN architecture instead of the previously proposed MarioGAN as it seems obvious that the latter would not easily be able to encode global symmetry. Preliminary tests also confirmed that the solutions of the baseline versions were visually similar.

%In both questions, an original version is tested against its adaptation, thus evaluating their relative performance. To provide a sense of their absolute performance, 
We add original levels from CCS as baselines (Original1 and Original2). The original levels are converted to the same level representation as described in Section~\ref{sec:pcgml-candycrush} in order to ensure comparability. Each generator produced $1,000$ levels for visual and computational analysis. For the survey, we select three levels from each dataset, which are the levels with minimal, median and maximal vertical symmetry score (defined in Section~\ref{sec:pcgml-computational-measures}), respectively. This is intended to ensure that some, but not all levels, adhere to the most common and very obvious type of global pattern, i.e.\ vertical symmetry.

%\begin{itemize}
%    \item mrf-local vs mrf-global vs. orig
%    \item standardGAN-horiz vs. standardGAN vs. orig\todo{still needs to train}
%    \item mrf-local-mirror vs. standardGAN-mirror vs. orig
%\end{itemize}

%\todo[inline]{mention that we also modify originals}

%---Results---
\section{Results}
\label{sec:results}
%-----
\subsection{Visual Analysis}
\label{sec:visual-analysis}
\begin{figure}
     \centering
     %\begin{subfigure}[b]{.32\columnwidth}
     %    \centering
     %    \includegraphics[width=\textwidth]{pics/levels_selected/mrf-local-146-game.png}
     %    \caption{LocalMRF min}
     %    \label{fig:visa}
     %\end{subfigure}
     %\hspace{0.1\columnwidth}
     \begin{subfigure}[b]{.32\columnwidth}
         \centering
         \includegraphics[width=\textwidth]{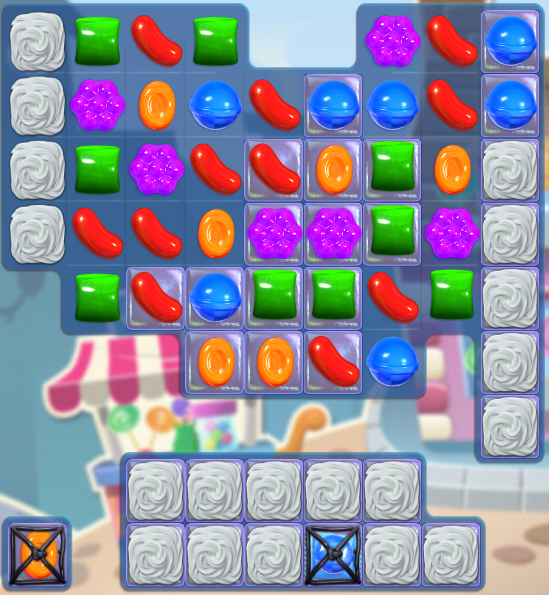}
         \caption{LocalMRF}
         \label{fig:visb}
     \end{subfigure}
    \hspace{0.1\columnwidth}
    % \begin{subfigure}[b]{.32\columnwidth}
    %     \centering
    %     \includegraphics[width=\textwidth]{pics/levels_selected/mrf-global-748-game.png}
    %     \caption{GlobalMRF min}
    %     \label{fig:visc}
    % \end{subfigure}
    % \hspace{0.1\columnwidth}
     \begin{subfigure}[b]{.32\columnwidth}
         \centering
         \includegraphics[width=\textwidth]{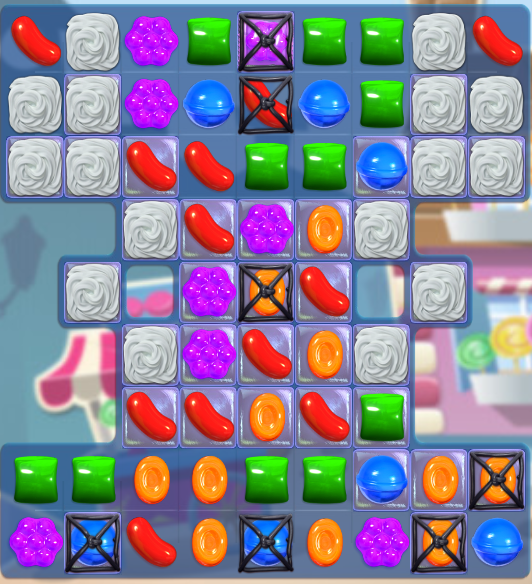}
         \caption{GlobalMRF}
         \label{fig:visd}
     \end{subfigure}
     \\
    % \begin{subfigure}[b]{.32\columnwidth}
    %     \centering
    %     \includegraphics[width=\textwidth]{pics/levels_selected/oneGAN-457-game.png}
    %     \caption{GlobalGAN min}
    %     \label{fig:vise}
    %\end{subfigure}
    %\hspace{0.1\columnwidth}
     \begin{subfigure}[b]{.32\columnwidth}
     \centering
         \includegraphics[width=\textwidth]{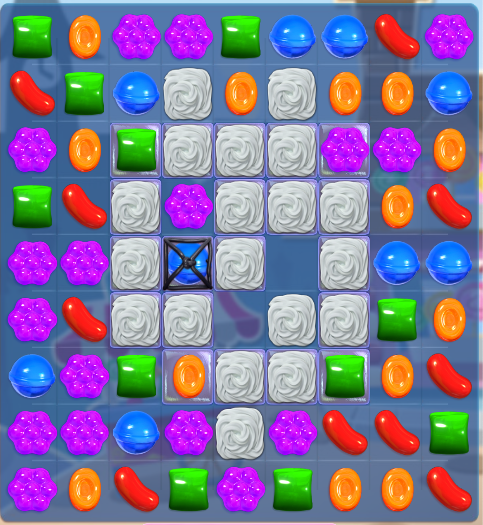}
         \caption{GlobalGAN}
         \label{fig:visf}
    \end{subfigure}
    \hspace{0.1\columnwidth}
    %\begin{subfigure}[b]{.32\columnwidth}
    %    \centering
    %        \includegraphics[width=\textwidth]{pics/levels_selected/oneGAN-horiz_58-game.png}
    %    \caption{\scriptsize{GlobalGAN-vert min}}
    %    \label{fig:visg}
    %\end{subfigure}
    %\hspace{0.1\columnwidth}
    \begin{subfigure}[b]{.33\columnwidth}
        \centering
        \includegraphics[width=\textwidth]{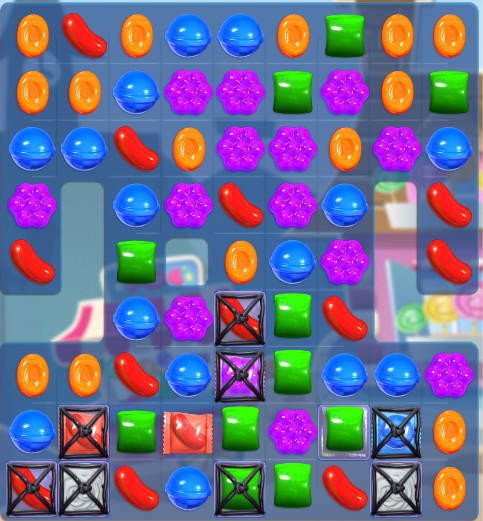}
        \caption{\footnotesize{GlobalGAN-vert}}
        \label{fig:vish}
    \end{subfigure}
    \\
    % \begin{subfigure}[b]{.32\columnwidth}
    %     \centering
    %     \includegraphics[width=\textwidth]{pics/levels_selected/orig-23-game.png}
    %     \caption{Original1 min}
    %     \label{fig:visi}
    % \end{subfigure}
    % \hspace{0.1\columnwidth}
     \begin{subfigure}[b]{.32\columnwidth}
         \centering
         \includegraphics[width=\textwidth]{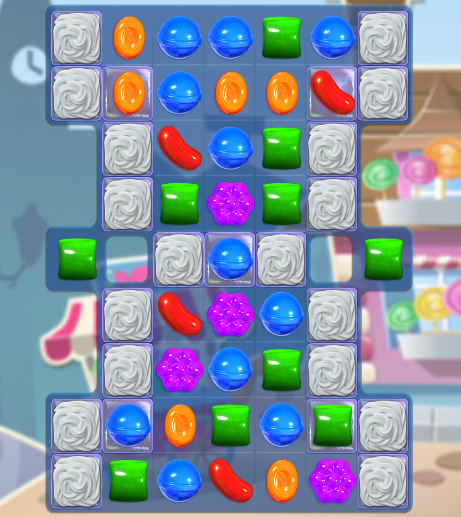}
         \caption{Original1}
         \label{fig:visj}
     \end{subfigure}
     %\\\
    %\begin{subfigure}[b]{.32\columnwidth}
    %     \centering
    %     \includegraphics[width=\textwidth]{pics/levels_selected/orig-154-game.png}
    %     \caption{Original2 min}
    %     \label{fig:visk}
    % \end{subfigure}
     \hspace{0.1\columnwidth}
     \begin{subfigure}[b]{.32\columnwidth}
         \centering
         \includegraphics[width=\textwidth]{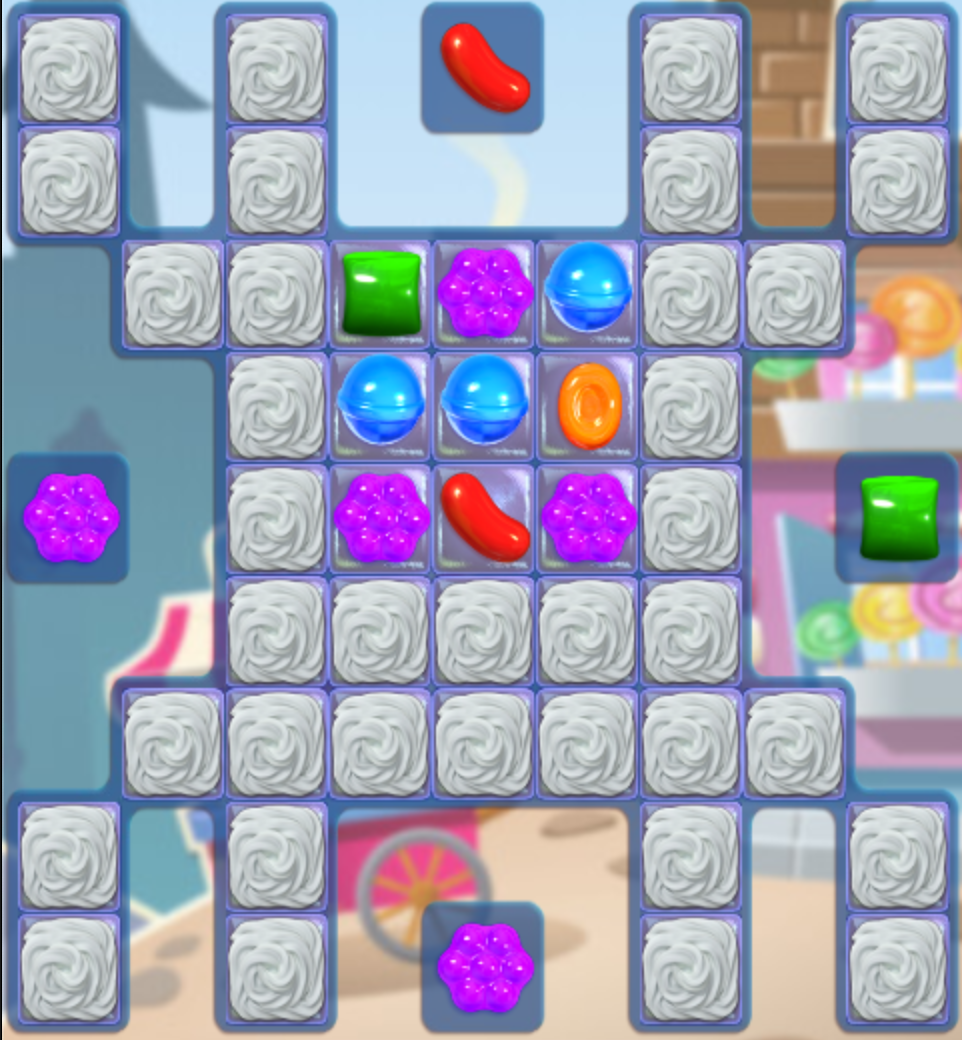}
         \caption{Original2}
         \label{fig:visl}
     \end{subfigure}
        \caption{Visualisation of most vertically symmetric CCS level from each set of levels.}
        \label{fig:vis}
\end{figure}

For each dataset  we visually analyse the level generators. 
%While we did an extensive analysis, here, 
Here, we specifically focus on the generators' ability to create vertically symmetric levels. We choose to analyse the best-case scenario here, thus cherry-picking the level with the maximum vertical symmetry score for each set of levels (Figure~\ref{fig:vis}).
%. The resulting levels are visualised in Figure~\ref{fig:vis}.

Exact local patterns are difficult to detect in all the levels. However, in all generated levels (Figures~\ref{fig:visb}-\ref{fig:vish}), locks and blockers appear in groups instead of on their own. This is a local pattern reflected in the original levels as well (Figures~\ref{fig:visj}-\ref{fig:visl}). This was expected as all methods have previously been shown to be able to reproduce local patterns (cf. Section~\ref{sec:background-patterns-existing}).

%Looking at the original levels with minimal vertical symmetry (Figures~\ref{fig:visi} and \ref{fig:visk}), it becomes apparent that they still contain some global structure. While they are not symmetric, they display different patterns on the left and right, each separately displays large patterns. The level in Figure~\ref{fig:visi} is actually horizontally symmetric and in the level depicted in Figure~\ref{fig:visk}, each half is vertically symmetric.

%In contrast, this does not appear to be true for the corresponding generated ``min'' levels in Figures \ref{fig:visa}, \ref{fig:visc}, \ref{fig:vise} and \ref{fig:visg}. They do not have any obvious global patterns.
The generated levels (Figures \ref{fig:visb}-\ref{fig:vish}) do show recognisable vertical symmetry. While the addition of information (RQ1: GlobalMRF vs. LocalMRF) seems to improve the generator's ability to generate symmetric levels recognisably, the difference seems to be less stark when enriching the data (RQ2: GlobalGAN vs. GlobalGAN-vert).

%As a first conclusion, 
%we can say from anecdotal evidence, 
The results suggest that the generators are able to capture some global patterns (namely vertical symmetry), but not consistently. No other global patterns were observed, but it is unclear whether this is due to the small sample size. %The adaptations seem to result in some improvements.

%--- Testing single column figures

\subsection{Computational Measures}

\begin{figure*}
     \centering
     \begin{subfigure}[b]{0.29\textwidth}
         \centering
         \includegraphics[width=\textwidth, page=1]{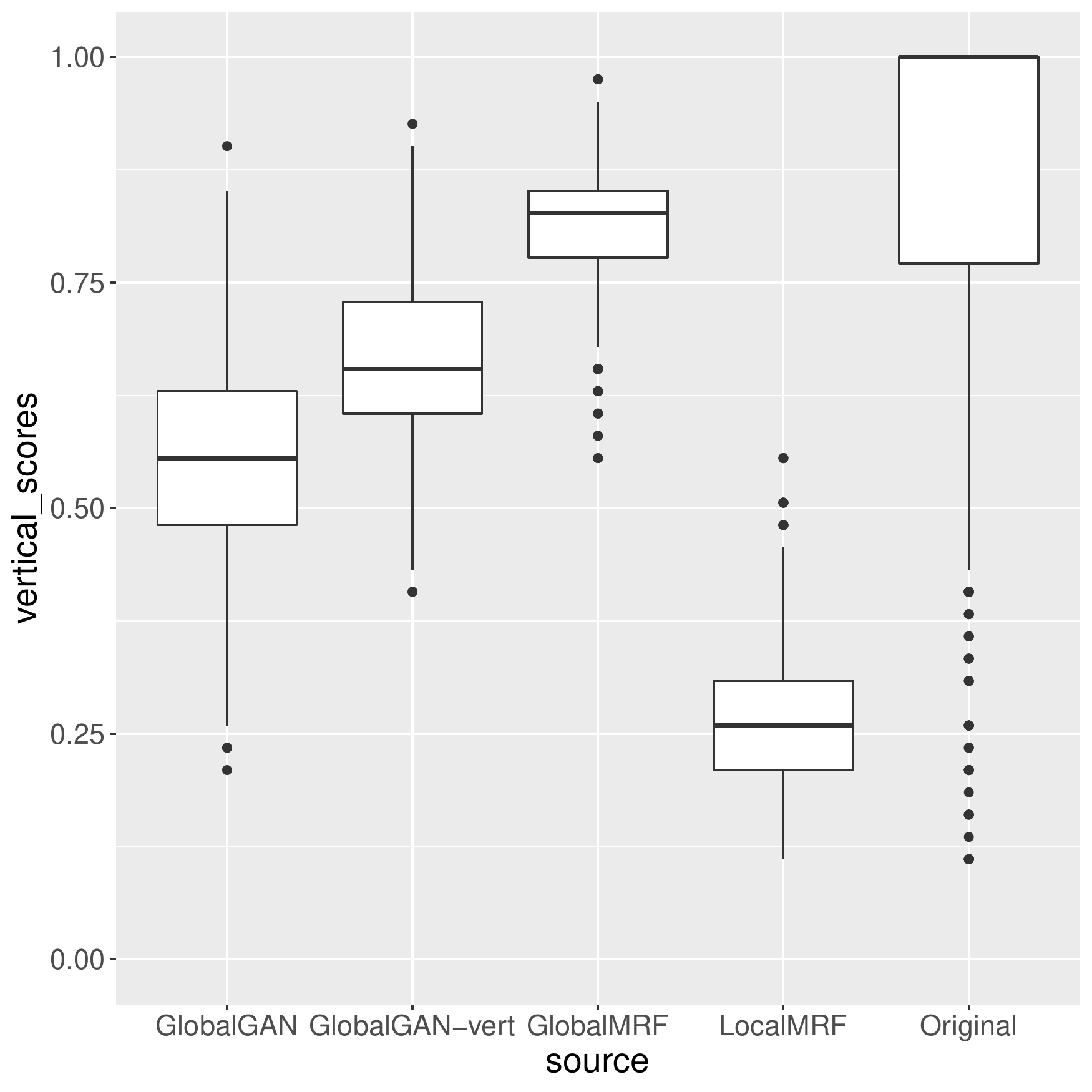}
         \caption{Vertical symmetry.}
         \label{fig:comp1}
     \end{subfigure}
     \hfill
     \begin{subfigure}[b]{0.29\textwidth}
         \centering
         \includegraphics[width=\textwidth, page=2]{pics/boxplots.pdf}
         \caption{Horizontal Symmetry}
         \label{fig:comp2}
     \end{subfigure}
     \hfill
     \begin{subfigure}[b]{0.29\textwidth}
         \centering
         \includegraphics[width=\textwidth, page=3]{pics/boxplots.pdf}
         \caption{Diagonal symmetry.}
         \label{fig:comp3}
    \end{subfigure}
        \caption{Boxplots of symmetry scores  of each set of levels. All generators are able to create more or less symmetric levels, with the GlobalMRF being able to create vertical and horizontal symmetric levels better than any other approach. The vertical symmetry scores of the GAN approach can be improved slightly by exclusively training on vertical levels. }
        \label{fig:comp}
\end{figure*}

As the above analysis is anecdotal, we seek to validate the results using computational measures, i.e symmetry scores as defined in Section~\ref{sec:pcgml-computational-measures}. To this end, we generated $1\,000$ levels with each generator and computed their symmetry scores. The resulting scores are shown in Figure~\ref{fig:comp}, together with the symmetry measures computed for the original levels.

Most of the original levels are vertically symmetric, with some apparent variety (Figure~\ref{fig:comp1}). This variety is even bigger across the other symmetry dimensions (Figures~\ref{fig:comp2} and \ref{fig:comp3}). The generators are able to learn from this data and create individual levels with reasonable symmetry scores. However, their symmetry score distributions are vastly different from the training level - showing much less variety and lower scores across the board. Adapting PCGML methods so they can also capture the variety of original levels along several dimensions would be an interesting direction for future research, for example by guiding the sampling of original levels appropriately.

Notice that LocalMRF has very low scores for horizontal and vertical symmetry, as is expected given the strictly local neighbourhood of the model. It is, however, slightly better in terms of diagonal symmetry. This could be a result of a cascading effect caused by sampling with the local neighbourhood, but further investigation is needed.%This could be because diagonally symmetric local patterns dictate more of the global patterns of the levels. In order to obtain a reliable explanation, however, further investigation is needed.

\paragraph*{RQ1}
Extending the neighbourhood with symmetric neighbours (GlobalMRF) noticeably increases the symmetry scores along the horizontal and vertical axes. Diagonal information was not added, resulting in a decreased diagonal symmetry score. This highlights a potential problem with this type of adaptation - only global patterns that are explicitly captured will be encouraged with this method, while others (diagonal symmetry in this case) could be negatively affected. We leave further exploration of this for future work.

\paragraph*{RQ2}
The GlobalGAN shows average scores across the board, and a definite improvement in vertical symmetry is achieved by training only on vertically symmetric levels (GlobalGAN-vert). The vertical scores also vary less, but still quite considerably. However, the score improvement is not major, which may be because most original levels are already vertically symmetric. The performance in other symmetry dimensions is not affected. This suggests that the vertically symmetric levels show sufficient variety to be able to capture other types of symmetry. If the resulting sample was too small or too homogeneous, we would likely see adverse effects. Further work is needed to confirm this conclusion in general.

%-------
\subsection{Survey Results}
\label{sec:pattern-aware-survey}

We obtained $60$ responses from King employees to our user study. We received answers from people working on various games, of which $63\%$ work on match-three games and almost $40\%$ work on CCS specifically. The participants have different roles in the company, with $14$ working in level design.

\begin{figure*}
     \centering
     \begin{subfigure}[b]{0.45\textwidth}
        \centering
        \includegraphics[page=1, width=\textwidth, trim={0.2cm 6cm 0cm 6.7cm},clip]{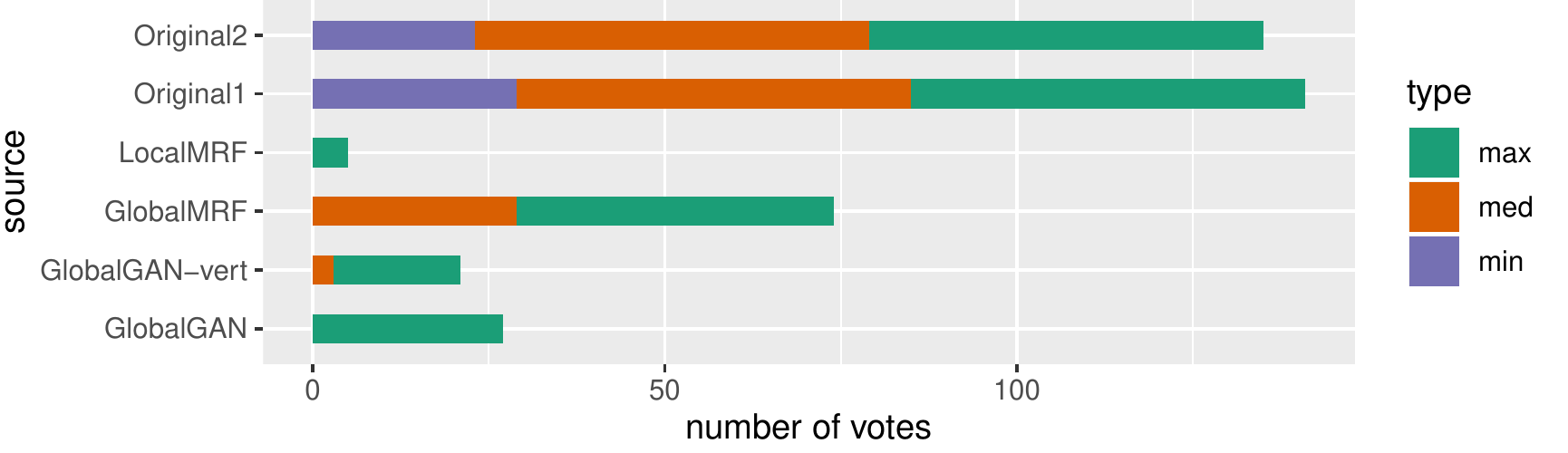}
        \caption{Indicate all levels that contain a global pattern.}
        \label{fig:global}
     \end{subfigure}
     \hfill
      \begin{subfigure}[b]{0.45\textwidth}
        \centering
         \includegraphics[page=3, width=\textwidth, trim={0.2cm 5.8cm 0cm     6.3cm},clip]{pics/output_plots.pdf}
        \caption{RQ1: Which level looks less like a Candy Crush level?}
        \label{fig:info_vis}
     \end{subfigure}
     \\
     \begin{subfigure}[b]{0.45\textwidth}
         \centering
        \includegraphics[page=2, width=\textwidth, trim={0.2cm 6cm 0cm 6.7cm},clip]{pics/output_plots.pdf}
        \caption{Indicate all levels that contain a local pattern.}
        \label{fig:local}
     \end{subfigure}
         \hfill
      \begin{subfigure}[b]{0.45\textwidth}
        \centering
         \includegraphics[page=5, width=\textwidth, trim={0.2cm 5.8cm 0cm     6.3cm},clip]{pics/output_plots.pdf}
        \caption{RQ2: Which level looks less like a Candy Crush level?}
        \label{fig:data_vis}
     \end{subfigure}
        \caption{Questionnaire results. (a, c) Participants were presented with three levels from
each method with the minimal, median and
maximal vertical symmetry score. GlobalMRF is better than any of the other methods but the original levels still contain more recognizable global and local patterns. The pairwise comparisons between the different methods (b, d) show that levels generated by GlobalMRF are more likely considered CCS levels than any of the other approaches but are still considered to look less like CCS levels than the original levels.
}
        \label{fig:survey_results}
\end{figure*}

The original levels have more recognisable global patterns according to the study. This aligns with our results both from the visual and computational analysis. The generators are mostly not able to produce global patterns (Figure~\ref{fig:global}). An outlier is GlobalMRF, which creates levels with a comparatively high amount of vertical symmetry (Figure~\ref{fig:comp1}), which is also reflected in the study results.

The distinction is far less clear when it comes to local patterns (Figure~\ref{fig:local}). Local patterns were recognised much less often than global ones across all levels. The original levels mostly still have an advantage, but GlobalMRF is a close third.

\begin{figure}
     %\centering
     %\begin{subfigure}[b]{0.4\columnwidth}
     %    \centering
     %    \includegraphics[width=\textwidth]{pics/levels_selected/ex4_abs.png}
     %    \caption{example1}
     %    \label{fig:valid1}
     %\end{subfigure}
     %\hspace{0.05\columnwidth}
     %\begin{subfigure}[b]{0.4\columnwidth}
     %    \centering
     %    \includegraphics[width=\textwidth]{pics/validation/examples-4-game.png}
     %    \caption{example1}
     %    \label{fig:valid2}
     %\end{subfigure}\\
     \begin{subfigure}[b]{.48\columnwidth}
         \centering
         \includegraphics[width=0.7\textwidth]{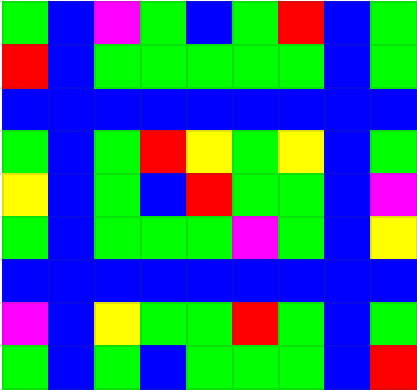}
         \caption{Abstract Representation}
         \label{fig:valid3}
    \end{subfigure}
    \hfill
    \begin{subfigure}[b]{.48\columnwidth}
        \centering
        \includegraphics[width=0.7\textwidth]{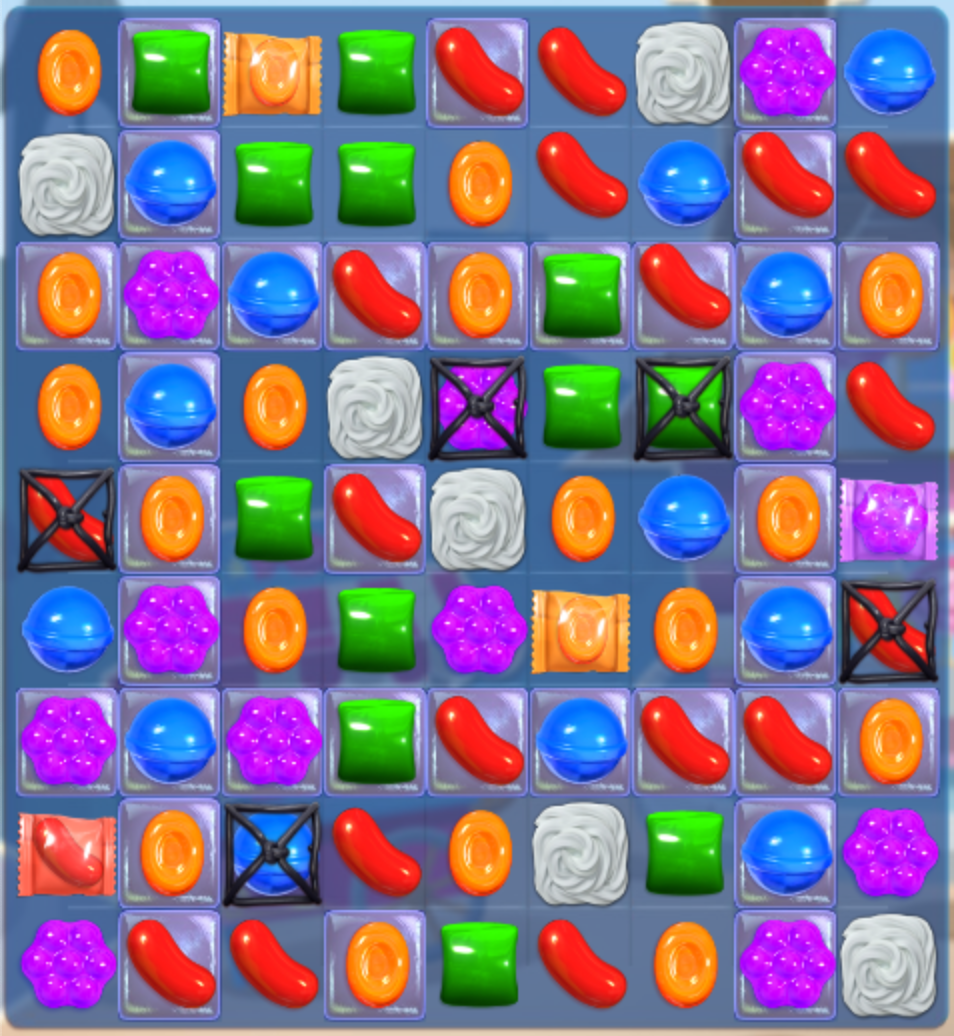}
        \caption{Visualisation in survey}
        \label{fig:valid4}
    \end{subfigure}
        \caption{Validation example from the questionnaire. While specific patterns are clearly visible when shown in color (a), they become less obvious when shown as game tiles (b).}
        \label{fig:valid}
\end{figure}

Note, however, that human perception is influenced by visual aspects such as colours. So in some cases, structural patterns may exist, but are not easily recognisable by a human. We tested this by adding levels to the questionnaire that were deliberately constructed to contain a specific pattern. % An example can be found in Figure~\ref{fig:valid}. %Even though the local pattern in Figure~\ref{fig:valid2} was designed to be obvious and aligned well with the example given in the questionnaire (cf. Figure~\ref{fig:ex1}), $11$ of the $60$ respondents did not recognise a local pattern. Further, 
For example, even though a global pattern is clearly identifiable in Figure~\ref{fig:valid3}, it is not very noticeable when visualised as a CCS level (Figure~\ref{fig:valid4}). As a consequence, only eight participants indicated that they saw a global pattern in the level.

%--- testing single column figures for space saving

%\begin{figure*}
%     \centering
%     \begin{subfigure}[b]{0.20\textwidth}
%         \centering
%         \includegraphics[width=\textwidth]{pics/levels_selected/ex4_abs.png}
%         \caption{example1}
%         \label{fig:valid1}
%     \end{subfigure}
%     \hfill
%     \begin{subfigure}[b]{0.20\textwidth}
%         \centering
%         \includegraphics[width=\textwidth]{pics/validation/examples-4-game.png}
%         \caption{example1}
%         \label{fig:valid2}
%     \end{subfigure}
%     \hfill
%     \begin{subfigure}[b]{0.20\textwidth}
%         \centering
%         \includegraphics[width=\textwidth]{pics/levels_selected/ex5_abs.png}
%         \caption{example1}
%         \label{fig:valid3}
%    \end{subfigure}
%    \hfill
%    \begin{subfigure}[b]{0.20\textwidth}
%        \centering
%        \includegraphics[width=\textwidth]{pics/validation/examples-5-game.png}
%        \caption{example1}
%        \label{fig:valid4}
%    \end{subfigure}
%        \caption{caption}
%        \label{fig:valid}
%\end{figure*}

We conclude that humans are not always able to explicitly identify patterns. For the evaluation of our research questions, we thus relied more on pairwise comparisons that only require intuition instead of an explicit detection of patterns.

%            G   L
%4  local	6	49
%5  global	8	8
%6  both	39	28

%Adding global information to the neighbourhood of the Markov Random Fields has demonstrably improved their visual likeness to Candy Crush levels (cf. figure \ref{fig:info_vis}). While the original levels still perform better
\paragraph*{RQ1}
In the pairwise comparison as depicted in Figure~\ref{fig:info_vis}, we can clearly see that most respondents considered the LocalMRF levels less likely to be CCS levels when compared with GlobalMRF. In both cases most respondents identify the generated level as dissimilar from CCS levels when compared with an original level. Still, based on this data, some of the respondents could see generated levels be part of the corpus of CCS levels, even when compared with originals. Levels from the augmented algorithm GlobalMRF were preferred more often. We conclude that adapting the MRF algorithm by adding symmetrical inputs improved its ability to create global patterns that make them recognisable as CCS levels.

\paragraph*{RQ2}
The pairwise comparison (Figure~\ref{fig:data_vis}) is also favourable for the adaptation here, GlobalGAN-vert when compared to the baseline algorithm GlobalGAN. The adaptation is detected as an atypical CCS level less often. We thus also deem this adaptation to be beneficial for creating recognisable global patterns.

%--- DISCUSSION----
\section{Discussion}
\label{sec:discussion}

Based on our visual, computational, and survey results we conclude that the implemented adaptations are indeed beneficial in improving the underlying algorithms' ability to produce recognisable global patterns.  
%This conclusion stems from three different types of analyses (visual, computational, survey) which all agreed in this regard. 
However, our main intention in this paper is not promote a specific adaptation, but to demonstrate untapped potential for improving existing PCGML methods specifically with regards to their ability to produce global patterns. This is a claim we can make, even if our exact results are not necessarily generalisable across applications and / or PCGML algorithms. We assume that adaptations suitable for GANs should also be suitable for other neural network-based approaches (Variational Autoencoders, Convolutional Neural Networks) and would expect similar results. Sequential models (Markov Chains, Long Short-Term Memory networks) probably require more specific adaptations (like GlobalMRF), because generation and modelling takes place in sequence. However, as shown in this paper, encoding domain knowledge can greatly improve the obtained results.

Our observations regarding the fact that GANs seem to not be able to consistently produce symmetric levels are interesting because one of the algorithm's most popular applications is generating faces which usually display imperfect vertical symmetry. We investigated whether a lack of consistent symmetric training signal in CCS could explain this by training on vertically symmetric levels exclusively. While we observed small improvements in this regard (Figure~\ref{fig:comp1}), this explanation evidently does not fully account for the lack of symmetry in the output. There are several other potential explanations, but one avenue that should be investigated more is how the discretisation of the level representation  (Section~\ref{sec:pcgml-candycrush}) affects the output. GANs have been applied to level generation in various tile-based games, % (e.g.\ Mario \cite{Volz2018} and Doom \cite{8848011}), 
all requiring such discretisation.

%------
%\subsubsection{Why do GANs not produce symmetry like in faces?}
%\label{pcgml-gans}
%Hypotheses:
%\begin{itemize}
%    \item Data for faces has consistent global structure (symmetry - always horizontal, eyes always in the upper area) - not true for CC. This way no strong enough training signal.
%    \item Complexity of genotype-phenotype mapping destroys global structure that is generated
%    \item approach only models local patterns (adjacent tiles), and local patterns do not automatically limit global ones
%    \item model is not complex enough to model all different types of patterns at the same time
%\end{itemize}

A more general observation based on our list of potential pattern-aware adaptations to PCGML algorithms in Section~\ref{sec:pattern-aware-adaptation}, is that they all require varying levels of domain knowledge. However, even domain experts have difficulty identifying patterns (cf. Section~\ref{sec:pattern-aware-survey}). It is thus important to consider that such required domain knowledge might not always be available when applying a PCGML algorithm. This result also highlights the importance of testing PCG approaches on different domains in order to be able to assess them independently of potential unconscious biases in the training data and / or algorithms. We are looking to confirm our results on other games where (1) non-exact patterns are important (like imperfectly symmetric faces) and that (2) have reasonable complexity so the application of PCGML is warranted.

Further,  several of the pattern-aware adaptations mentioned in Section~\ref{sec:pattern-aware-adaptation} resemble some techniques from other machine learning domains. 
%SEB: pretty old refernce. not sure this is the current way people do this so commented out for now
%In time series analysis, for example, it is common to include seasonal models and general trends in the models \cite{timeseries}.
It therefore seems worthwhile to investigate further which other adaptations are made in related fields that operate on data with patterns.

Additionally, it is important to note that evaluating generated content in PCG computationally is still an open challenge. This is especially true if the measure is intended to target something that is not easily formalised and / or relies on human perception. Even visual patterns, as targeted in this paper, proved difficult to evaluate despite the fact that humans do not need to play the game to identify them.

Lastly, this paper focused only on visual patterns, i.e.\ patterns observable from the representation of the generated content. However, ultimately, the goal of PCGML is to create content that not only looks, but also plays similarly to the given set of levels. It is thus important to study further how visual patterns relate to gameplay patterns in different game genres. We also plan to conduct a deeper investigation of the expressive range of different PCGML methods.

%---CONCLUSION---
\section{Conclusions}
\label{sec:conclusion}

The results in this paper suggest that match-three games such as Candy Crush Saga provide a challenge for current state-of-the-art PCGML methods due to the fact that they often include larger scale (global) structural patterns. Several different ways to improve existing methods were outlined, and some investigated further. The implemented adaptations proved to be successful in our experiments and related research questions should be pursued further. However, significant work remains to be done, especially for generalising, formalising, extending and rigorously testing pattern-aware adaptations of existing PCGML algorithms. 
In addition, we plan to further explore characterisation and similarity measures to pick up on global and local patterns while minimising the required domain knowledge. %Promising avenues seem to be similarity measures based on tile-pattern distributions (\todo{ref kldiv}) or based on clustering results (\todo{ref?}). 
It would also be interesting to investigate more how the discrete representation of the generated content affects the performance of the respective algorithm, especially for GANs. We also plan to study whether levels with similar visual patterns also result in similar gameplay.

\bibliographystyle{IEEEtran}
\bibliography{pcgml}

\end{document}